\documentclass[conference]{IEEEtran}
\IEEEoverridecommandlockouts
\usepackage{cite}
\usepackage{amsmath,amssymb,amsfonts}
\usepackage{algorithmic}
\usepackage{graphicx}
\usepackage{textcomp}
\usepackage{xcolor}
\usepackage{multirow}
\usepackage{subcaption}
\usepackage[a4paper, total={184mm,239mm}]{geometry}

\def\BibTeX{{\rm B\kern-.05em{\sc i\kern-.025em b}\kern-.08em
    T\kern-.1667em\lower.7ex\hbox{E}\kern-.125emX}}
\graphicspath{{./figures}}
\begin{document}
\title{
    Enhancing Neural Network Robustness Against Fault Injection Through Non-linear
    Weight Transformations
	\thanks{
		This research is based on results obtained from a project, JPNP16007,
		commissioned by the New Energy and Industrial Technology Development
		Organization (NEDO).
	}
}

\makeatletter
\newcommand{\linebreakand}{\end{@IEEEauthorhalign} \hfill\mbox{}
	\par
	\mbox{}\hfill\begin{@IEEEauthorhalign}}
\author{
	\IEEEauthorblockN{1\textsuperscript{st} Ninnart Fuengfusin}
	\IEEEauthorblockA{\textit{Graduate School of Life Science and Systems} \\
		\textit{Kyushu Institute of Technology}\\
		Kitakyushu, Japan \\
		ninnart@brain.kyutech.ac.jp}
	\and

	\IEEEauthorblockN{2\textsuperscript{nd} Hakaru Tamukoh}
	\IEEEauthorblockA{\textit{Graduate School of Life Science and Systems} \\
		\textit{Research Center for Neuromorphic AI Hardware}\\
		\textit{Kyushu Institute of Technology}\\
		Kitakyushu, Japan\\
		tamukoh@brain.kyutech.ac.jp}
}
\makeatother
\maketitle

\begin{abstract}
Deploying deep neural networks (DNNs) in real-world environments poses challenges due to
faults that can manifest in physical hardware from radiation, aging, and temperature
fluctuations. To address this, previous works have focused on protecting DNNs via
activation range restriction using clipped ReLU and finding the optimal clipping
threshold. However, this work instead focuses on constraining DNN weights by applying
saturated activation functions (SAFs): Tanh, Arctan, and others. SAFs prevent faults
from causing DNN weights to become excessively large, which can lead to model failure.
These methods not only enhance the robustness of DNNs against fault injections but also
improve DNN performance by a small margin. Before deployment, DNNs are trained with
weights constrained by SAFs. During deployment, the weights without applied SAF are
written to mediums with faults. When read, weights with faults are applied with SAFs and
are used for inference. We demonstrate our proposed method across three datasets
(CIFAR10, CIFAR100, ImageNet 2012) and across three datatypes (32-bit floating point
(FP32), 16-bit floating point, and 8-bit fixed point). We show that our method enables
FP32 ResNet18 with ImageNet 2012 to operate at a bit-error rate of 0.00001 with minor
accuracy loss, while without the proposed method, the FP32 DNN only produces random
guesses. Furthermore, to accelerate the training process, we demonstrate that an
ImageNet 2012 pre-trained ResNet18 can be adapted to SAF by training for a few epochs
with a slight improvement in Top-1 accuracy while still ensuring robustness against
fault injection.
\end{abstract}


\section{Introduction}
Deep neural networks (DNNs) have been gaining attention due to their ability to
achieve the state-of-the-art performance across various tasks \cite{yu2024inceptionnext,
touvron2023llama, wang2024yolov10}. Furthermore, with the introduction of large-language
models (LLMs), the ability of DNNs to generalize across tasks has dramatically improved
\cite{brown2020language}. However, this often comes with a trade-off in terms of their
size and computational overheads.

Serving large-scale DNNs requires a number of hardware, from memory to processors. This
increases the overall likelihood of faults manifesting in physical hardware, as these
faults can be caused by temperature fluctuations \cite{cha2017defect}, aging
\cite{dixit2021silent}, write errors \cite{nozaki2019recent}, and other factors. These
faults may cause bit-flips in DNN parameters and can easily trigger models failures
\cite{hong2019terminal}, which can be catastrophic in safety-critical applications.

To address this issue, several research methods have been proposed. One of these is
activation-restriction-based methods, which are designed based on bounded activation
functions, such as ReLU6 and Tanh. These activation functions can suppress
high-intensity activation values that are likely to be caused by faults and improve the
overall numerical stability of DNNs under faults \cite{hong2019terminal}. Further work
in this field focuses on ReLU-like activation functions that include a threshold to
map high-intensity positive activation values to zero \cite{hoang2020ft, chen2021low,
ghavami2022fitact, mousavi2024proact}.

Instead of focusing on activations, this work focuses on DNN weights, which are more
sensitive to faults compared to activations \cite{reagen2018ares}. Since weights are
reused across inputs, it is more crucial to protect them. Before deployment, our DNNs
are trained with weights constrained by saturated activation function (SAF) that maps
outputs to a bounded range. During deployment, the weights, without the applied SAF, are
written to fault-prone mediums. When in use, weights with faults are applied with SAF to
mitigate large deviations caused by faults. An overview of our proposed method is shown
in Fig. \ref{fig:overview}.

\begin{figure}[htbp]
	\centerline{\includegraphics[scale=0.3]{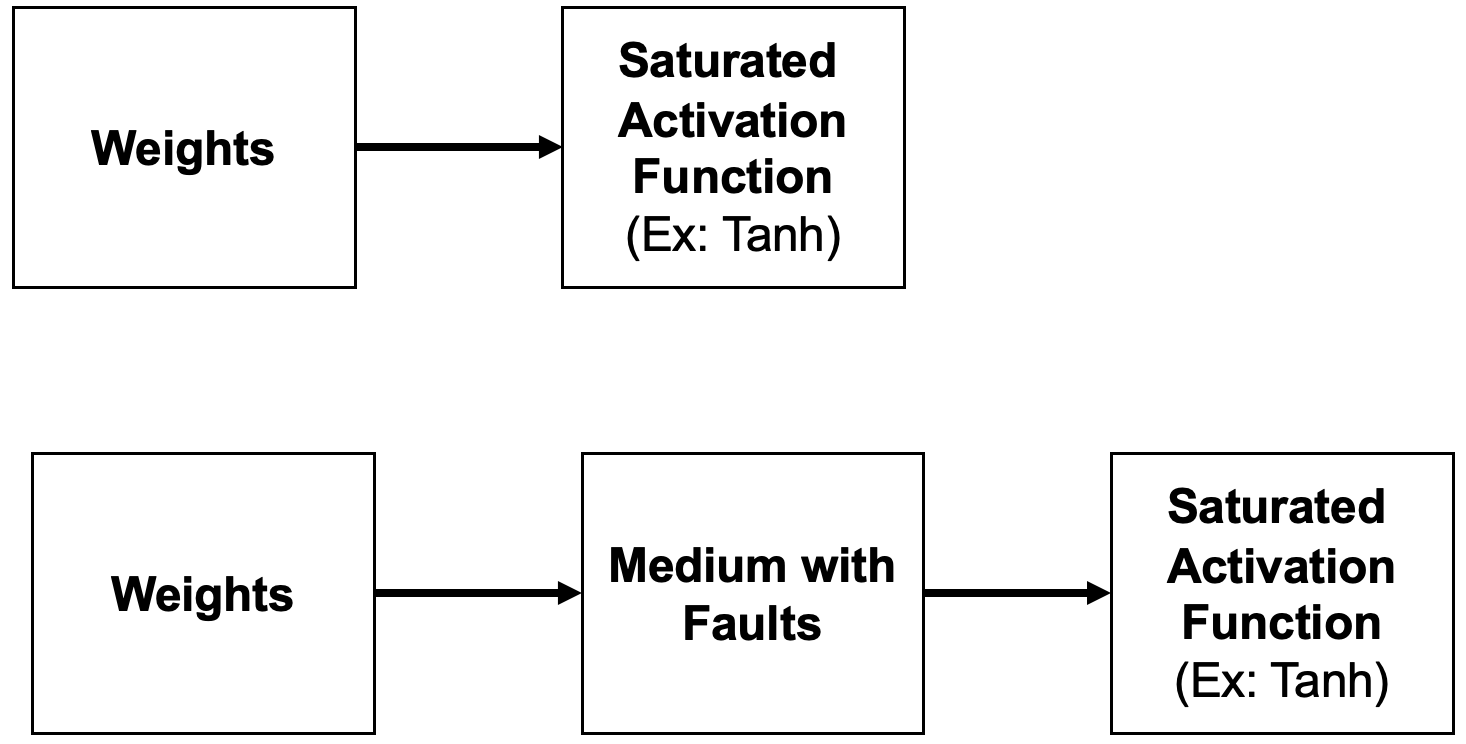}}
	\caption{
        Overview of our proposed method. \textbf{Top:} During training, weights are
        passed through to an SAF, allowing the DNN to aware of the SAF. \textbf{Bottom:}
        During deployment to fault-prone mediums, weights without SAF are stored within
        these mediums. When use, weights with faults are applied with the SAF, which
        suppresses large deviations caused by faults.
	}
	\label{fig:overview}
\end{figure}

The use of non-linear activation functions on weights was proposed by
\cite{cakaj2023weight} as a regularization technique. Our work expands on this concept
to harden DNN weights against fault injections. Instead of Arctan that
\cite{cakaj2023weight} based on, we explore several additional activation functions:
Tanh, modified Tanh, and Softsign. To accelerate the SAF adaptation process, we
demonstrate that our methods can be fine-tuned using \texttt{torchvision}
\cite{torchvision2016} ImageNet 2012 \cite{deng2009imagenet} pre-trained weights, rather
than training from scratch. This reduces the number of training epochs from hundreds to
only a few epochs, with a slight improvement in top-1 accuracy.

We demonstrate that our method enables 32-bit floating point (FP32) or 16-bit floating
point (FP16) models, which are highly sensitive to bit-flips, to operate at a bit-error
rate (BER) of $10^{-5}$ with only minor performance loss, without any bells and
whistles. In contrast, baseline FP32 and FP16 DNNs without SAFs produce random guesses
under BER $= 10 ^{-5}$. Furthermore, our method generalizes to 8-bit fixed point (Q2.5,
or 1-bit sign, 2 bit integer, and 5-bit fraction) models, enhancing their robustness
against fault injections.

Our main contributions are listed as follows:

\begin{itemize}
    \item We propose a method to enhance DNN tolerance against bit-flips by applying
        SAFs to weights. Our method enables FP32 and FP16 models to operate at a BER of
        $10^{-5}$ without any additional modifications. Additionally, our method enhances
        the robustness of Q2.5 models against bit-flips.
    \item We demonstrate that our method generalizes across three datasets: CIFAR10,
        CIFAR100, and ImageNet 2012, and across three datatypes: FP32, FP16, and Q2.5.
    \item We show that our method can be accelerated by fine-tuning from commonly
        available ImageNet 2012 pre-trained weights to adapt to SAFs.
\end{itemize}

\section{Related Works}
In this section, we describe two research directions related to our method. The first is
the use of non-linear activation functions applied to DNN weights, and the second is the
use of bounded activation functions to enhance DNN tolerance to bit-flips.

The utilization of non-linear activation functions with DNN weights was introduced by
the weight compounder \cite{cakaj2023weight}. The weight compounder applies an
Arctan-based non-linear activation function to weights, regularizing DNNs by
discouraging weights that are close to zero and large weights, which indicate signs of
dead weights and over-fitting, respectively.

For the second research direction, the use of bounded activation functions to enhance
DNN tolerance against bit-flips was proposed by \cite{hong2019terminal}. This work
demonstrates that bounded activation functions, such as ReLU6 and Tanh, can suppress
high-intensity activations from propagating across DNNs. In the same direction,
\cite{hoang2020ft} introduces a clipped ReLU, which is based on ReLU with an additional
threshold to map high-intensity positive activation values to zero. To determine the
optimal clipping threshold, several methods have been proposed, ranging from profiling
values \cite{chen2021low}, fine-tuning \cite{hoang2020ft}, to training-base approaches
\cite{mousavi2024proact}.

Our work extends the weight compounder \cite{cakaj2023weight} by applying SAFs to harden
DNNs against fault injections, and further introduces additional SAF functions,
including Tanh, modified Tanh, and Softsign. While our approach is similar to
activation-restriction-based methods in that both methods map input values into a
bounded range, our method focuses on weight restrictions, as weights are more sensitive
to faults compared to activations \cite{reagen2018ares}. Another distinction is that our
method does not rely on ReLU activation functions, which cannot be directly applied to
DNN weights.

The downside of activation-restriction-based methods is their computational complexity,
which increases with the number of input data. In contrast, our method depends on the
number of weights, which remains constant regardless of the number of input data. This
limitation becomes more significant when serving DNNs in large-scale applications, as
the number of input data grows with the number of users.

\section{Proposed Method}
Given the $i$-th layer weight as $W_{i}$ and the output activation from the $(i - 1)$-th
layer as $a_{i - 1}$. The affine transformation followed by a non-linear activation,
commonly used in feed-forward DNNs, is defined in \eqref{eq:basic}. Here, $\sigma$ is a
non-linear activation function, and $b_{i}$ is the $i$-th layer bias term.

\begin{equation}
	a_{i} = \sigma(W_{i}a_{i - 1} + b_{i})
	\label{eq:basic}
\end{equation}

During training, the key distinction between our proposed method and \eqref{eq:basic} is
that our method applies $\tau$, or SAFs, to the weights, as shown in
\eqref{eq:proposed}.

\begin{equation}
	a_{i} = \sigma(\tau(W_{i})a_{i - 1} + b_{i})
	\label{eq:proposed}
\end{equation}

SAF is a non-linear activation function that saturates high-intensity input values into
a bounded range. Different $\tau$ functions penalize high-intensity values in varying
ways, as shown in Fig. \ref{fig:acts}. The faster the output values reach the saturation
point, the stronger the penalization of high-intensity input values will be.

This penalization effect can be controlled to an extent by scaling the input values
before applying SAF. To demonstrate this, we introduce the modified Tanh. The modified
Tanh is a Tanh function with an additional hyper-parameter $c$, defined as $tanh(cx)$,
where $tanh$ is the Tanh function, $x$ is the input value, and $c$ is a hyper-parameter
that controls how quickly the output value converges to the saturated value. Since Tanh
strongly penalizes high-intensity values compared to other $\tau$ functions, we
introduce the modified Tanh with $c = 0.5$ (Tanh0.5) to relax the penalization effect.

\begin{figure}[htbp]
	\centerline{\includegraphics[scale=0.45]{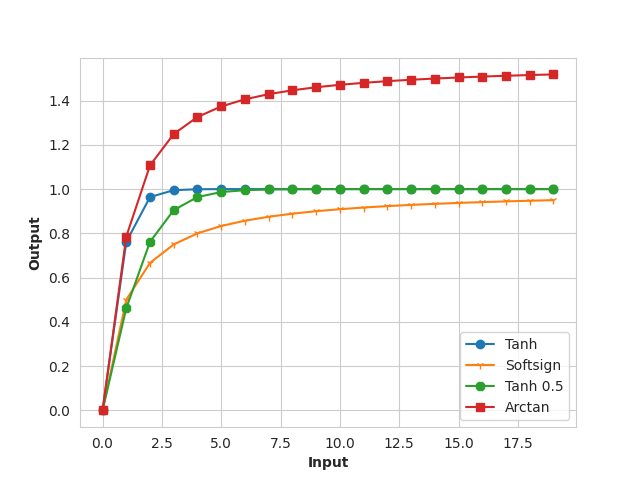}}
	\caption{
        Relation between positive input values and outputs of different SAFs.
	}
	\label{fig:acts}
\end{figure}

During deployment to fault-prone mediums, bit-flips may occur in $W_{i}$. Let $f$
represent a function that injects bit-flips with a bit-error rate (BER). Thus, the
weights with faults, $\hat{W}_{i}$ are denoted as shown in \eqref{eq:fault}.

\begin{equation}
	\hat{W}_{i} = f(W_{i}, BER)
	\label{eq:fault}
\end{equation}

To suppress the effects from bit-flips, $\tau$ is used to constrain weights within a
bounded range, as shown in \eqref{eq:inference}. This ensures the numerical stability of
DNNs and prevents large deviations caused by faults. Since our weights are trained to be
aware of SAFs, applying SAFs during inference does not result in a loss in model
performance.

\begin{equation}
	a_{i} = \sigma(\tau(\hat{W}_{i})a_{i - 1} + b_{i})
	\label{eq:inference}
\end{equation}

Hence, the overhead of our proposed method during deployment is limited to applying
$\tau$ to the weights.

\section{Experimental Results and Discussion}
In this section, we conduct experiments across three datasets: CIFAR10, CIFAR100, and
ImageNet 2012 \cite{deng2009imagenet}.

Since faults occur randomly, we simulate them by performing Monte Carlo simulations,
randomly injecting bit-flips into weights with a probability of BER $= 10^{-5}$. For
CIFAR10 and CIFAR100, we conducted 100 rounds of simulations, while for ImageNet 2012,
we performed 10 rounds. The average and standard deviation of the top-1 test accuracy
are reported in the format of Mean $\pm$ Standard Deviation.

We reported the results across three datatypes: FP32, FP16, and Q2.5. Since the default
datatype for DNN is FP32, we reported it as is. For FP16 and Q2.5 datatypes, we
converted from FP32 weights. Note that all DNN operations were performed in the FP32
datatype, with the exceptions of bit-flips, which were injected into the weights with
their respective datatypes.

\subsection{CIFAR10 and CIFAR100}
For CIFAR10 and CIFAR100, ResNet20 \cite{he2016deep} was used as our base model. In each
experiment, we trained ResNet20 with SAFs applied to the weights from scratch. Our
training settings were as follows: we used a batch size of 128, a stochastic gradient
descent optimizer with a momentum of 0.9, and an initial learning rate of 0.1. The
models were trained for 200 epochs with a weight decay of $10^{-3}$. The learning rate
was decayed using a cosine learning rate scheduler.

We reported the results of models using four SAFs: Tanh, Tanh0.5, Softsign, and Arctan.
For the baseline without SAFs, we reported the results with SAF as "None". The
experimental results are shown in Table \ref{tab:cifar10} for CIFAR10 dataset and Table
\ref{tab:cifar100} for CIFAR100 dataset.

\begin{table}[htbp]
    \caption{
        ResNet20 test accuracy on the CIFAR10 dataset before fault injections (Top-1 \%)
        and after fault injections (After Top-1 \%).
    }
	\begin{center}
		\begin{tabular}{|c|c|c|c|c|}
			\hline
			\textbf{Datatype} & \textbf{SAF} & \textbf{Top-1 (\%)} & \textbf{After Top-1 (\%)} \\
            \hline
            \multirow{5}{*}{FP32} & None & 92.34 & 38.61 $\pm$ 37.31 \\
            \cline{2-4}
            & Tanh & 92.07 & 89.25 $\pm$ 7.92 \\
            \cline{2-4}
            & Tanh0.5 & \textbf{92.6} & 80.68 $\pm$ 16.91 \\
            \cline{2-4}
            & Softsign & 92.5 & \textbf{90.64 $\pm$ 3.24} \\
            \cline{2-4}
            & Arctan & 92.56 & 83.45 $\pm$ 13.83 \\
            \hline

            \multirow{5}{*}{FP16} & None & 92.33 & 34.67 $\pm$ 35.35 \\
            \cline{2-4}
            & Tanh & 92.10 & 89.99 $\pm$ 4.62 \\
            \cline{2-4}
            & Tanh0.5 & \textbf{92.59} & 82.02 $\pm$ 15.72 \\
            \cline{2-4}
            & Softsign & 92.51 & \textbf{90.13 $\pm$ 4.12} \\
            \cline{2-4}
            & Arctan & 92.55 & 83.71 $\pm$ 15.30 \\
            \hline

            \multirow{5}{*}{Q2.5} & None & 91.08 & 60.12 $\pm$ 25.91 \\
            \cline{2-4}
            & Tanh & 90.47 & 85.02 $\pm$ 10.25 \\
            \cline{2-4}
            & Tanh0.5 & 90.67 & 81.36 $\pm$ 11.04 \\
            \cline{2-4}
            & Softsign & \textbf{91.21} & \textbf{89.68 $\pm$  2.9}1 \\
            \cline{2-4}
            & Arctan & 90.38 & 85.19 $\pm$  8.68 \\
            \hline
		\end{tabular}
		\label{tab:cifar10}
	\end{center}
\end{table}

For CIFAR10 dataset, we observed that Tanh0.5 provides the best performance without
fault injections for both FP32 and FP16 datatypes. For the Q2.5 datatype, Softsign
offers a better alternative. However, after fault injections, Softsign delivers the best
test accuracies across all datatypes. Applying SAFs to the weights significantly
mitigates the reduction in test accuracy caused by fault injections.

In the best case, for FP32, the top-1 accuracy reduction is reduced from 53.73 without
SAFs to 1.86 with Softsign. For FP16, the top-1 accuracy reduction is reduced from 57.66
without SAFs to 2.38 with Softsign. For Q2.5, the top-1 accuracy reduction is reduced
from 30.96 without SAFs to 1.53 with Softsign. As demonstrated in Table
\ref{tab:cifar10}, Softsign also provides the lowest standard deviation across all
datatypes.

Compared to Q2.5 models, FP32 and FP16 models are more sensitive to bit-flips, as
shown by both lower mean and higher standard deviation in top-1 accuracy. This
sensitivity arises from their high dynamic range, which allows for dramatic changes in
weight values after bit-flips \cite{hong2019terminal}. However, by applying SAFs, we can
ensure that FP32 and FP16 models operate at a BER of $10^{-5}$ with only minor loss in
model performance.

Furthermore, we conducted experiments with ResNet20 on the CIFAR10 dataset across
different BER values, as shown in Fig. \ref{fig:ber}. From Fig. \ref{fig:ber}, our SAFs
enhanced the robustness of DNNs across BER values ranging from $10^{-6}$ to $10^{-5}$,
while models without SAFs struggled to operate at a BER of $10^{-6}$. However, we
observed that none of SAFs could operate at a BER of $10^{-4}$.

\begin{figure*}[htbp]
	\centering
	\begin{subfigure}{0.32\textwidth}
		\includegraphics[width=\textwidth]{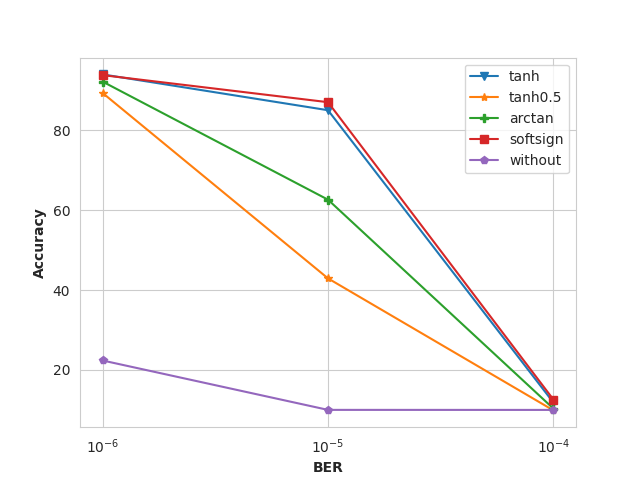}
		\caption{32-bit floating point}
		\label{fig:ber1}
	\end{subfigure}
	\hfill
	\begin{subfigure}{0.32\textwidth}
		\includegraphics[width=\textwidth]{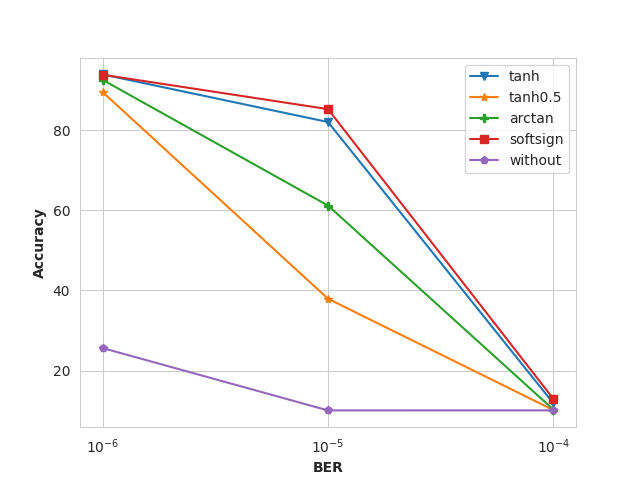}
		\caption{16-bit floating point}
		\label{fig:ber2}
	\end{subfigure}
	\hfill
	\begin{subfigure}{0.32\textwidth}
		\includegraphics[width=\textwidth]{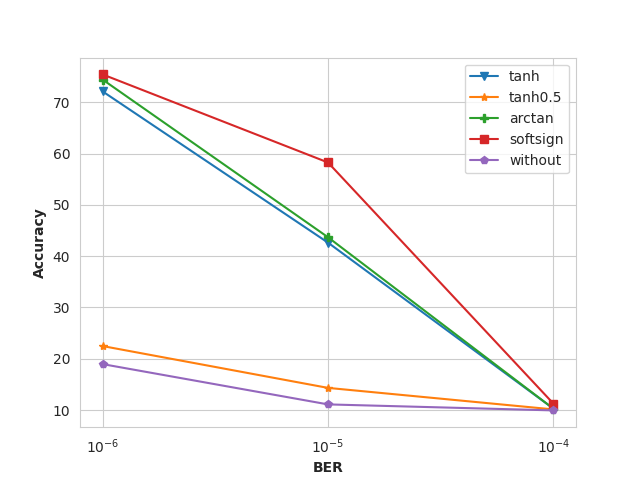}
		\caption{8-bit fixed point (Q2.5)}
		\label{fig:ber3}
	\end{subfigure}
	\caption{Test accuracy of ResNet20 on CIFAR10 across different BER values.}
	\label{fig:ber}
\end{figure*}

For the CIFAR100 dataset, using the same experimental settings as for CIFAR10, the
experimental results are shown in Table \ref{tab:cifar100}.

\begin{table}[htbp]
    \caption{
        ResNet20 test accuracy on the CIFAR100 dataset before fault injections (Top-1 \%)
        and after fault injections (After Top-1 \%).
    }
	\begin{center}
		\begin{tabular}{|c|c|c|c|c|}
			\hline
			\textbf{Datatype} & \textbf{SAF} & \textbf{Top-1 (\%)} & \textbf{After Top-1 (\%)} \\
            \hline
            \multirow{5}{*}{FP32} & None & 68.77 & 19.37 $\pm$ 28.34 \\
            \cline{2-4}
            & Tanh & 68.65 & \textbf{66.10 $\pm$ 3.93} \\
            \cline{2-4}
            & Tanh0.5 & \textbf{69.44} & 56.00 $\pm$ 15.81 \\
            \cline{2-4}
            & Softsign & 68.55 & 65.12 $\pm$ 6.45 \\
            \cline{2-4}
            & Arctan & 68.62 & 60.30 $\pm$ 11.54\\
            \hline

            \multirow{5}{*}{FP16} & None & 68.73 & 19.40 $\pm$ 28.54 \\
            \cline{2-4}
            & Tanh & 68.66 & \textbf{66.52 $\pm$ 2.53} \\
            \cline{2-4}
            & Tanh0.5 & \textbf{69.46} & 56.90 $\pm$ 14.81\\
            \cline{2-4}
            & Softsign & 68.49 & 65.73 $\pm$ 3.15 \\
            \cline{2-4}
            & Arctan & 68.61 & 62.27 $\pm$ 8.41\\
            \hline

            \multirow{5}{*}{Q2.5} & None & 65.92 & 37.47 $\pm$ 20.68 \\
            \cline{2-4}
            & Tanh & 65.72 & 61.10 $\pm$ 5.08 \\
            \cline{2-4}
            & Tanh0.5 & \textbf{66.56} & 56.50 $\pm$ 10.12\\
            \cline{2-4}
            & Softsign & 66.04 &  \textbf{63.80 $\pm$ 4.02} \\
            \cline{2-4}
            & Arctan & 66.32 & 61.46 $\pm$ 5.00 \\
            \hline
		\end{tabular}
		\label{tab:cifar100}
	\end{center}
\end{table}

From Table \ref{tab:cifar100}, we observed that Tanh0.5 provides the best performance
without fault injections across all datatypes. However, after fault injections, Tanh
delivers the best top-1 accuracy for FP32 and FP16. For Q2.5 datatype, Softsign is the
better alternative. Similar to the CIFAR10 dataset, applying SAFs significantly
mitigates the overall reduction of top-1 accuracy reduction after bit-flips.

For FP32, the top-1 accuracy reduction is reduced from 49.40 without SAFs to 2.55 with
Tanh. For FP16, the top-1 accuracy reduction decreases from 49.33 without SAFs to 2.14
with Tanh. For Q2.5, the top-1 accuracy reduction is reduced from 28.45 without SAFs to
2.24 with Softsign. Similar to Softsign in the CIFAR10 dataset, Tanh provides the
lowest standard deviation across all datatypes.

The best candidate SAFs for both the CIFAR10 and CIFAR100 datasets are Softsign and
Tanh. However, Softsign and Tanh are oppposite in terms of penalizing high-intensity
values. Softsign is more relaxed, while Tanh is more penalizing. Tanh0.5 lies between
Softsign and Tanh but performs worse relative to both. Another distinction between
Softsign and Tanh is the curve of their SAFs, which we hypothesize may contribute to
their robustness against bit-flips.

\subsection{ImageNet 2012}
So far, our methods have been trained from scratch to allow the DNN to be aware of SAF.
However, this experiment demonstrates that the training process for our method can be
accelerated by using pre-trained weights.

To demonstrate this, we used ResNet18 \cite{he2016deep} with a pre-trained weight from
\texttt{torchvision} \cite{torchvision2016}. We fine-tuned this model for 5 epochs with
a batch size of 128, a initial learning rate of $10^{-5}$ with a cosine learning rate
scheduler, and the AdamW optimizer \cite{loshchilov2017decoupled} with a weight decay of
$10^{-3}$

In this experiment, Tanh was selected as one of the best SAF candidate from the CIFAR10
and CIFAR100 sections. We could not directly fine-tune with SoftSign, as we hypothesized
that SoftSign might cause numerical instability when training with pre-trained weights.
If SAF is set to None, it indicates that we directly utilized the \texttt{torchvision}
pre-trained weights for inferences without applying SAFs. The experimental results are
shown in Table \ref{tab:before}.

\begin{table}[htbp]
    \caption{
        ResNet18 test accuracy on ImageNet 2012 before fault injections (Top-1 \%)
        and after fault injections (After Top-1 \%).
    }
	\begin{center}
		\begin{tabular}{|c|c|c|c|}
			\hline
			\textbf{Datatype} & \textbf{SAF} & \textbf{Top-1 (\%)} & \textbf{After Top-1 (\%)} \\
            \hline
            \multirow{2}{*}{FP32} & None & 69.76 & 0.10 $\pm$ 0.00 \\
            \cline{2-4}
            & Tanh & \textbf{70.33} & \textbf{67.26 $\pm$ 1.01} \\
            \hline
            \multirow{2}{*}{FP16} & None & 69.75  & 0.10 $\pm$ 0.01 \\
            \cline{2-4}
            & Tanh & \textbf{70.31} & \textbf{67.48 $\pm$ 0.81}\\
            \hline
            \multirow{2}{*}{Q2.5} & None & 66.87 & 48.04 $\pm$ 5.90 \\
            \cline{2-4}
            & Tanh & \textbf{67.05} & \textbf{62.58 $\pm$ 1.42} \\
			\hline
		\end{tabular}
		\label{tab:before}
	\end{center}
\end{table}

From Table \ref{tab:before}, we observed that by fine-tuning ResNet18 with Tanh, the
top-1 accuracy improved by 0.57. Furthermore, after fault injections, the top-1 accuracy
of the baseline models dropped to random guesses for both FP32 and FP16. However, since
Q2.5 is more robust to bit-flips, its top-1 accuracy only reduced by 18.83. With our
proposed method using Tanh as the SAF, the top-1 accuracy dropped by only 3.07 for FP32,
2.83 for FP16, and 4.47 for Q2.5.

We also evaluated ResNet18 on the ImageNet 2012 dataset across different BER values, as
shown in Fig. \ref{fig:ber_imagenet}.

\begin{figure}[htbp]
	\centerline{\includegraphics[scale=0.45]{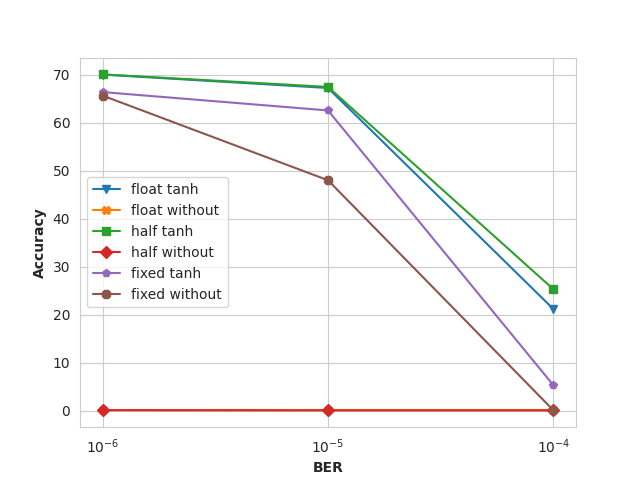}}
	\caption{
        ResNet18 test accuracy on ImageNet 2012 across different BER values.
	}
	\label{fig:ber_imagenet}
\end{figure}

From Fig. \ref{fig:ber_imagenet}, our SAFs significantly enhance the robustness of DNNs
across BER values from $10^{-6}$ to $10^{-5}$. However, while still better than random
guesses, our SAFs experience top-1 accuracy reduction of more than half when operating
at a BER of $10^{-4}$.

\section{Conclusion}
We propose a method to enhance the DNN tolerance against bit-flips by applying SAFs to
weights. To let DNN aware of SAFs, this can be achieved by either training the models
from scratch or fine-tuning from commonly pre-trained weights. Before deployment, the
weights without SAFs are written to fault-prone mediums. When read, weights with faults
are applied with SAFs to suppress large deviations caused by the faults. As the result,
the overheads of our method is minimal, involving only to apply SAF to weights. We
demonstrate that our proposed method enables models to operate at a BER of $10^{-5}$
with minor loss in test accuracy from fault injections across three datasets and three
datatypes.

\bibliographystyle{IEEEtran}
\bibliography{reference}
\end{document}